\documentclass[10pt,twocolumn,letterpaper]{article}

\usepackage{config/iccv}
\usepackage{config/iccv23}
\acrodef{voe}[VoE]{Violation of Expectation}
\acrodef{method}[\texttt{XPL}]{eXplanation-based Physics Learner}

\newcommand{\dataset}{\texttt{X-VoE}\xspace}
\iccvfinalcopy
\ificcvfinal\fi

\begin{document}

\title{\dataset: Measuring eXplanatory Violation of Expectation in Physical Events}

\author{\fontsize{11}{11}\selectfont{}Bo Dai$^{1,2}$, Linge Wang$^{3}$, Baoxiong Jia$^{2}$, Zeyu Zhang$^{2}$, Song-Chun Zhu$^{1,2,3}$, Chi Zhang$^{2,\textrm{\Letter}}$, Yixin Zhu$^{4,\textrm{\Letter}}$
    \vspace{0.5em}\\
    \hspace{-12pt}\begin{tabular}{r l}
        \small\url{https://github.com/daibopku/X-VoE}                                 &
        \small$\textrm{\Letter}$\,\,\texttt{zhangchi@bigai.ai,\,yixin.zhu@pku.edu.cn}   \\
        \small$^{1}$ School of Intelligence Science and Technology, Peking University &
        \small$^{2}$ Beijing Institute for General Artificial Intelligence              \\
        \small$^{3}$ Department of Automation, Tsinghua University                    &
        \small$^{4}$ Institute for Artificial Intelligence, Peking University
    \end{tabular}
}
\maketitle
\ificcvfinal\thispagestyle{empty}\fi

\begin{abstract}
    Intuitive physics is pivotal for human understanding of the physical world, enabling prediction and interpretation of events even in infancy. Nonetheless, replicating this level of intuitive physics in artificial intelligence (AI) remains a formidable challenge. This study introduces \dataset, a comprehensive benchmark dataset, to assess AI agents' grasp of intuitive physics. Built on the developmental psychology-rooted \ac{voe} paradigm, \dataset establishes a higher bar for the explanatory capacities of intuitive physics models. Each \ac{voe} scenario within \dataset encompasses three distinct settings, probing models' comprehension of events and their underlying explanations. Beyond model evaluation, we present an explanation-based learning system that captures physics dynamics and infers occluded object states solely from visual sequences, without explicit occlusion labels. Experimental outcomes highlight our model's alignment with human commonsense when tested against \dataset. A remarkable feature is our model's ability to visually expound \ac{voe} events by reconstructing concealed scenes. Concluding, we discuss the findings' implications and outline future research directions. Through \dataset, we catalyze the advancement of AI endowed with human-like intuitive physics capabilities.
\end{abstract}

\begin{figure}[t!]
    \centering
    \includegraphics[width=\linewidth]{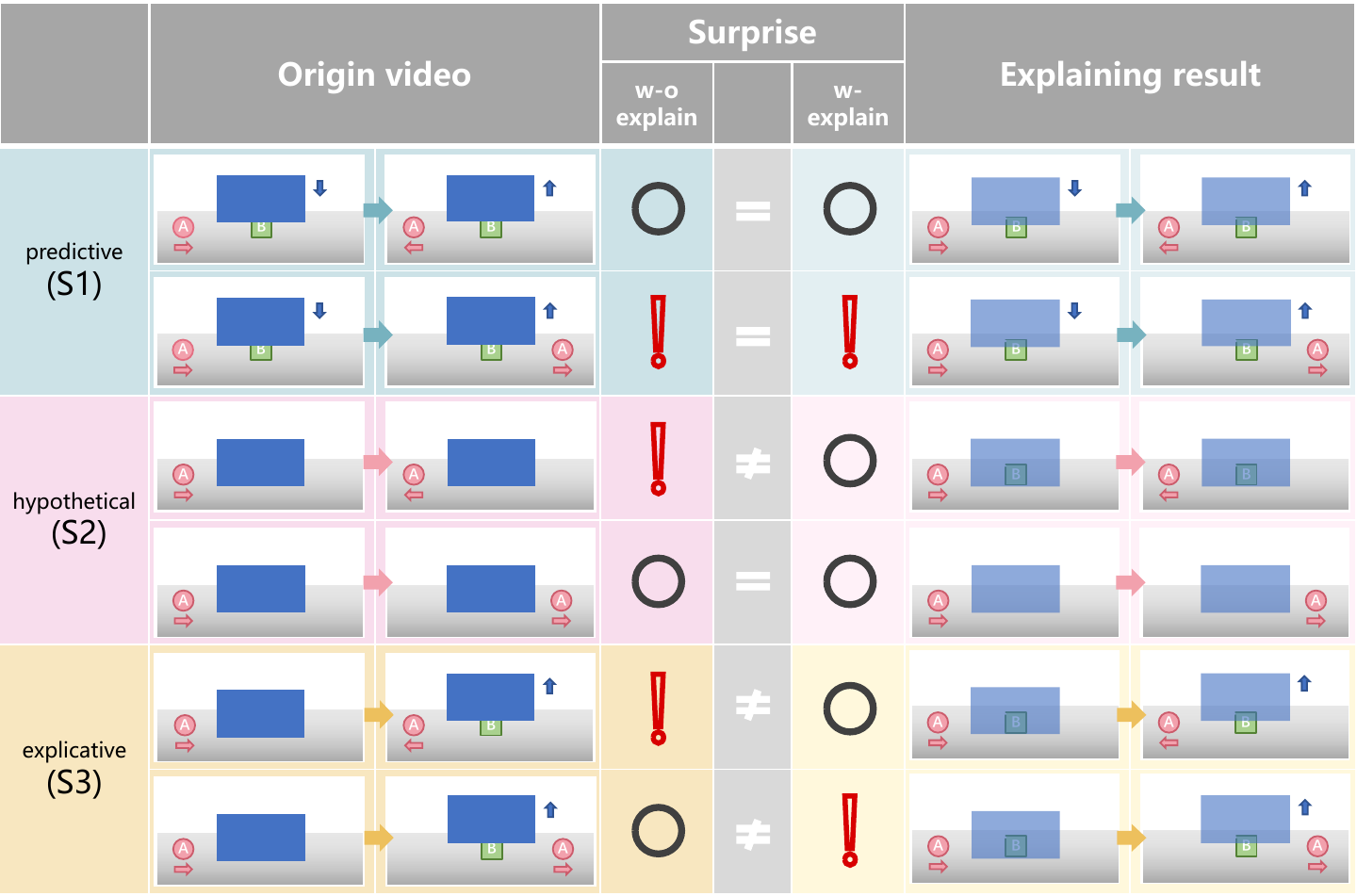}
    \caption{\textbf{Evaluation settings in the ball blocking exemplar scenario of \dataset.} The explanation video illustrates potential hidden dynamics. Circles denote no surprise, and exclamation marks indicate surprise. In the predictive setup (S1), a solvable pair is presented without requiring explanation: predicting observed entities' dynamics suffices to reason about the outcome. In the hypothetical setup (S2), perceiving the direction of outgoing balls might lead to surprise, yet alternate explanations exist---\eg, a hidden blocker behind the wall causing ball rebound. However, a random agent's scores show negligible disparity, necessitating the explicative setup (S3) to discern surprises, demanding explanatory ability absent in predictive-only or random agents.}
    \label{fig:explain}
\end{figure}

\section{Introduction}
Humans possess a profound understanding of the physical world, enabling them to predict the outcomes of physical interactions and events~\cite{battaglia2013simulation}. From infancy, humans demonstrate intuitive physics, comprehending actions and consequences even in unfamiliar scenarios. For the machine learning community, the challenge lies in emulating this level of intuitive physics understanding. This study introduces \dataset, a comprehensive benchmark dataset designed to assess and push the limits of AI agents' intuitive physics comprehension.

The notion of intuitive physics, observed even in young infants, has been foundational in cognitive science and developmental psychology~\cite{spelke2007core}. Infants show surprise when physical events violate their expectations, indicating an understanding of fundamental physical principles~\cite{baillargeon1985object}. Explanation-based learning has been proposed as a mechanism contributing to the development and refinement of intuitive physics understanding~\cite{baillargeon2017explanation}. However, recent advances in this field have primarily resulted in predictive models, lacking the explanatory capacity and falling short of capturing even infant-level intuitive physics comprehension~\cite{piloto2022intuitive,smith2019modeling}.

Central to our work is the \acf{voe} paradigm, widely employed in psychological studies to evaluate infants' intuitive physics understanding~\cite{baillargeon1994physical,baillargeon1985object}. In this paradigm, participants exhibit surprise, indicated by prolonged attention, when exposed to events that either follow or violate intuitive physics laws. Inspired by the effectiveness of this paradigm, we adopt it to evaluate AI agents' intuitive physics comprehension. In each trial, models encounter experiments adhering to or contravening intuitive physics laws. Models succeed in the VoE test if they display high surprise scores for physics-violating experiments and lower scores for compliant ones.

Existing works within the machine learning and computer vision community have embraced the \ac{voe} paradigm~\cite{dasgupta2021benchmark,piloto2022intuitive,riochet2020intphys,smith2019modeling,weihs2022benchmarking}. However, most of these efforts primarily focus on predictive abilities, disregarding the explanatory component~\cite{aguiar2002developments,piloto2018probing,piloto2022intuitive,riochet2020intphys,smith2019modeling,stahl2015observing}. This perspective neglects the fundamental aspect of \ac{voe}---the act of explaining observed events. In psychological studies, human participants express surprise not at the moment a physics-violating event occurs, but upon learning of its outcome. This observation underscores the significance of explanation within \ac{voe}.

\begin{figure*}[t!]
    \centering
    \includegraphics[width=\linewidth]{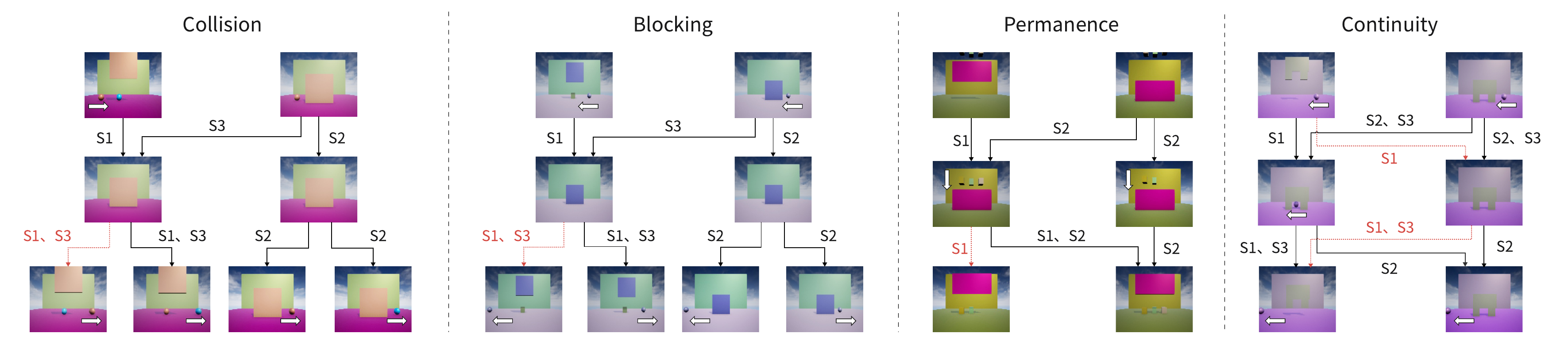}
    \caption{\textbf{Testing scenarios in \dataset: ball collision, blocking, object permanence, and object continuity.} Within each scenario, frames in a testing video are linked by the same setup identification number (\eg, S1). Black links denote non-surprising videos, while red links indicate surprising ones. Notably, certain videos require explanation to become non-surprising. For example, in the right S2 branch of the object permanence scenario, three cubes on the floor become non-surprising due to preceding observation of two cubes dropping, suggesting a hidden cube behind the wall.}
    \label{fig:test}
\end{figure*}

Motivated by these insights, we introduce \dataset, an intuitive physics evaluation dataset designed specifically to incorporate explanation within \ac{voe}. Distinct from previous efforts that concentrated on predictive scenarios, our dataset encompasses setups that require explaining observed events in diverse \ac{voe} situations. We establish three \ac{voe} settings for each of the four scenarios: ball collision, blocking, object permanence, and continuity (see \cref{fig:test}). Each scenario features predictive, hypothetical, and explicative setups. Notably, the three setups within the ball-blocking scenario distinguish explanatory agents from predictive and random ones.

Furthermore, we propose the \ac{method} model to emulate the explanation-based \ac{voe} process, inspired by findings in human studies~\cite{baillargeon1994physical,baillargeon2017explanation}. While \ac{method} is adaptable to diverse deep architectures, we specifically build it upon PLATO~\cite{piloto2022intuitive} due to its robust performance. Our model incorporates three self-supervised modules: perception for image encoding, Transformer reasoning for occluded object prediction, and dynamic reasoning for simulating physical dynamics. Importantly, our model introduces a reasoning sub-component to update representations of occluded objects, akin to infants' explanation-based learning when confronted with unexpected outcomes~\cite{baillargeon1994physical}.

In summary, our work makes three significant contributions:
\begin{itemize}[leftmargin=*,noitemsep,nolistsep]
    \item Introduction of \dataset, a comprehensive intuitive physics evaluation dataset that challenges AI agents not only in predictive capabilities but also in their capacity to explain. The dataset covers four distinct scenarios, each with predictive, hypothetical, and explicative setups. This allows for a more comprehensive assessment of intuitive physics understanding within \ac{voe}.
    \item Proposition of the \ac{method} model, enhancing existing approaches with an explanatory module that improves \ac{voe} evaluation. Our model comprises three modules---perception, reasoning, and dynamics learning---for holistic comprehension and simulation of physical dynamics.
    \item Experimental demonstration of \ac{method}'s enhanced performance in alignment with human commonsense compared to other baselines in \dataset. Additionally, \ac{method} offers insights into hidden factors, as depicted in \cref{fig:explain}.
\end{itemize}

\section{Related work}

\paragraph{Intuitive physics}

Intuitive physics forms a cornerstone of human cognition, enabling rapid and accurate predictions about moving object trajectories~\cite{kubricht2017intuitive}. To evaluate machine understanding in this realm, benchmark datasets have emerged, often focusing on predicting future states~\cite{battaglia2013simulation,chang2016compositional,lerer2016learning,wu2017learning,chen2020grounding} or inferring object properties~\cite{liang2016inferring,liang2018tracking,sanborn2013reconciling}. These methods predominantly gauge model performance by comparing generated predictions to ground truth.

More recently, the \acf{voe} paradigm has garnered attention within the machine learning and computer vision community~\cite{dasgupta2021benchmark,piloto2022intuitive,riochet2020intphys,smith2019modeling,weihs2022benchmarking}. Rooted in developmental psychology, the \ac{voe} paradigm quantifies model surprise when presented with events that challenge intuitive physics laws. This perspective provides an alternative angle for assessing intuitive physics understanding. Notably, the IntPhys dataset~\cite{riochet2020intphys} pioneered this \ac{voe}-based benchmarking approach. ADEPT~\cite{smith2019modeling} introduced a model combining re-rendering and object tracking. PLATO~\cite{piloto2022intuitive} decomposed the learning process into perception and dynamics prediction. Differing from conventional intuitive physics learning, the \ac{voe} paradigm does not rely on absolute ground truth. Instead, it hinges on relative measures of surprise, akin to developmental studies that assume higher responses indicate increased surprise. This emphasizes the role of explanation in \ac{voe}, as demonstrated in \cref{fig:explain}. In contrast to prior works that often neglected this vital component, our \dataset includes scenarios that demand both traditional prediction-based understanding and explanation-based comprehension. Additionally, we propose an explanation-enhanced physics learner, \ac{method}, which achieves improved performance and interpretability by incorporating explanations.

\paragraph{Video prediction}

The challenge of comprehending videos and making plausible predictions of future states from current observations has been a longstanding problem within computer vision~\cite{babaeizadeh2017stochastic,lotter2016deep,mathieu2015deep}, closely connected to the \ac{voe} paradigm. Solving \ac{voe} problems frequently involves predicting future frames for inference and evaluation. However, this prediction task is intricate due to the inherent complexity of modeling real-world dynamics and conditional image synthesis~\cite{tian2021good,weissenborn2019scaling}. Within the computer vision community, various architectures have been explored to address these challenges and enhance the quality of generated images~\cite{tian2021good,weissenborn2019scaling}. The task is further complicated by the need to model relationships between frames, leading to approaches that integrate spatial transformations over time~\cite{finn2016unsupervised,liu2017video,reda2018sdc}. Disentanglement of motion and content has also been pursued~\cite{denton2017unsupervised,hsieh2018learning,liu2021emergence,villegas2017decomposing}. More recent efforts involve learning physics-based dynamics from videos and reasoning about unknown factors~\cite{guen2020disentangling}. Within \dataset, we assess the performance of these video prediction models as baseline methods.

\paragraph{Object-centric dynamics}

The ``vision-as-inverse-graphics'' framework and the versatility of physics simulation have led to models based on physics simulation, which offer notable advantages in terms of accuracy and generality~\cite{chang2016compositional,riochet2020occlusion}. However, these models are often heavily reliant on specific physics engines, limiting their flexibility. In response, recent works have leveraged graph neural networks and object-centric representations to mitigate this dependence~\cite{piloto2022intuitive,watters2017visual}. By abstracting irrelevant signals and focusing on objects, these models establish a tighter mapping between visual inputs and physics engines. Further, some models can directly simulate real physics engines~\cite{battaglia2013simulation,ding2021dynamic,wu2017learning}. These object-centric dynamics models have demonstrated the ability to capture intricate dynamics. Our approach in \dataset aligns with this framework, using object-centric representations for downstream computation and reasoning.

\section{Generating \dataset}

Our \dataset dataset encompasses four distinct scenarios, covering ball collision, ball blocking, object permanence, and object continuity. To evaluate various intuitive physics principles, each scenario, except object permanence, comprises three distinct settings: predictive, hypothetical, and explicative, as illustrated in \cref{fig:test}. Within each setting, we create 1,000 procedurally generated scene pairs using Unreal Engine 4. Importantly, \dataset primarily serves as a test suite for evaluating intuitive physics understanding, with no constraints on model training data.

\subsection{Testing data}\label{sec:test_data}

We generate testing videos that span four key aspects of object dynamics: ball collision, ball blocking, object permanence, and object continuity. Refer to \cref{fig:test} for a visual overview.

\paragraph{Collision}

In this scenario, a ball traverses the scene, while an occlusion wall is positioned centrally. In the predictive setting (S1), we design a scenario where a ball of differing color but identical mass stands behind a wall. The incoming ball collides with this hidden ball, resulting in the incoming ball coming to a halt and the concealed ball continuing its trajectory. To introduce \ac{voe} effects, we enable the incoming ball to pass through the hidden ball. In the hypothetical setting (S2), we create a scene featuring a central wall concealing objects behind it. An incoming ball enters the scene from the left and rolls behind the wall. In some cases, an additional ball appears to pass through the wall, while in others, the incoming ball does so. This distinction hinges on whether an unseen ball is situated behind the wall. The explicative setting (S3) closely mirrors the hypothetical setting, but we lift the wall to reveal the concealed scene's contents.

\begin{figure*}[t!]
    \centering
    \includegraphics[width=\linewidth]{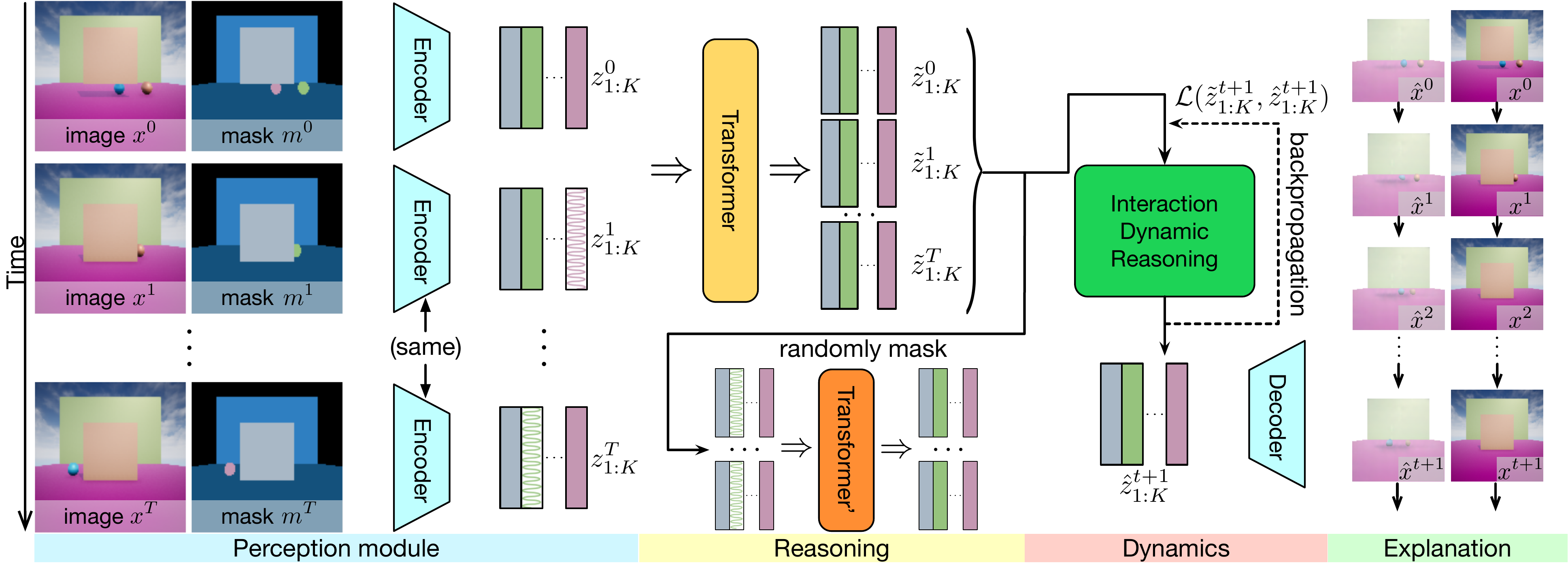}
    \caption{\textbf{Overview of the \ac{method} model for explanation-based physics learning.} The model comprises three key modules: (i) the perception module, responsible for extracting object-centric representation from RGBD videos and segmentation masks; (ii) the reasoning module, utilizing two Transformer networks to infer representations of occluded objects; (iii) the dynamics module, which acquires intuitive physical knowledge and refines reasoning outcomes to align with intuitive physics. Additionally, the inferred object representation can be visualized using the decoder from the perception module, offering a \textbf{visual explanation} of events occurring behind the wall. Wavy curves indicate masking. Refer to the text for comprehensive details.}
    \label{fig:pipeline}
\end{figure*}

\paragraph{Blocking}

The blocking scenario is conceptually similar to the collision scenario, substituting the hidden ball with a stationary cube. The impact of the incoming ball causes it to rebound upon collision with the cube.

\paragraph{Object permanence}

Drawing inspiration from developmental psychology literature, we recreate a scenario involving cubes falling to the ground and becoming occluded by a wall. In the predictive setting (S1), we devise a case where a wall descends to an initially vacant ground, followed by three cubes falling behind the wall. To elicit \ac{voe} effects, we raise the wall, revealing fewer than three objects. In the hypothetical setting (S2), the scenario begins with a wall positioned centrally, obscuring objects behind it. Three or two cubes fall behind the wall. When the wall is lifted, the scene consistently features three cubes, even when only two cubes initially fell. This reflects the possibility of one cube being hidden behind the wall from the outset.

\paragraph{Object continuity}

Motivated by psychology studies~\cite{aguiar2002developments}, we introduce a wall with a lower-half window. This setup allows a ball to traverse the scene from one side to the other. The ball becomes occluded when behind the wall, emerges through the window, disappears, and subsequently reappears from the opposite end. The three distinct settings mirror the collision and blocking scenarios. The differentiation between plausible and implausible scenes revolves around whether the ball remains visible upon passing through the window. In the predictive setting (S1), all relevant information is presented at the video's outset and conclusion, negating the presence of hidden objects. In the hypothetical setting (S2), information is deliberately withheld from the video's start and finish, necessitating the model's performance to align with infants~\cite{aguiar2002developments}, which involves explaining the existence of two balls. In the explicative setting (S3), the wall is lifted, verifying the absence of an additional ball behind the wall.

\subsection{Training data}

\begin{figure*}[t!]
    \centering
    \includegraphics[width=\linewidth]{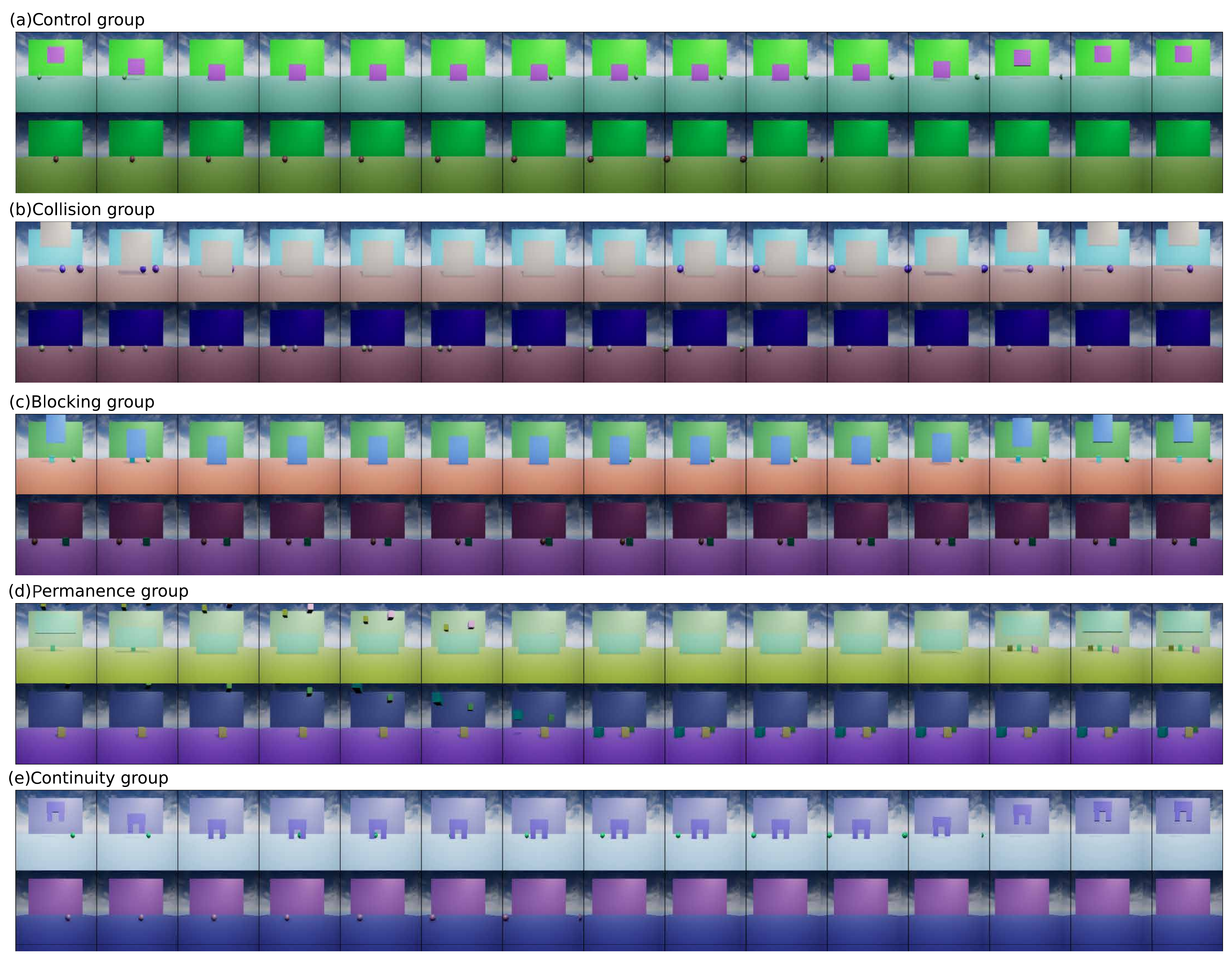}
    \caption{\textbf{Training scenarios for \dataset.} The timeline progresses from left to right, where each row represents the control, collision, blocking, object permanence, and object continuity groups from top to bottom. Please refer to \cref{sec_sup:train_data} for additional details.}
    \label{fig:train}
\end{figure*}

Though we do not impose constraints on the training data, for this study, we generate data adhering to the same structure as the test scenarios but without \ac{voe} effects. As shown in \cref{fig:train}, the training set consists of 100,000 procedurally generated scenes, closely mirroring the scale used for training PLATO~\cite{piloto2022intuitive}. During training, we exclusively present videos following intuitive physics laws, raising the wall at the beginning and end of each video. This approach reduces reasoning complexity, simulating the developmental process where only non-surprising physical events are observed. Consequently, models must unsupervisedly learn from video sequences depicting ordinary scenes, developing intuitive physics understanding necessary for \ac{voe}. Furthermore, for the collision and blocking scenarios, we create videos depicting balls passing through walls without collision or obstruction, demonstrating the unimpeded path behind the wall as shown in \cref{fig:train}(a). We also generate scenes similar to the previously described settings but devoid of occlusion walls.

\section{\texorpdfstring{\acf{method}}{}}

\subsection{Framework}

Our proposed \acf{method} model draws inspiration from developmental psychology theories concerning infancy. As depicted in \cref{fig:pipeline}, the \ac{method} model comprises three key components: (1) a perception module responsible for extracting object-centric representations to facilitate downstream processing, (2) a reasoning module tasked with inferring occluded object states by considering both spatial and temporal contexts, and (3) a dynamics module designed to acquire physical insights and evaluate inference outcomes for occluded objects.

\paragraph{Perception}

The perception module is designed to process input RGBD video sequences, represented as $\langle x^0, x^1, ..., x^T \rangle$, alongside their corresponding segmentation masks, denoted as $\langle m^0, m^1, ..., m^T \rangle$. The masks are generated using a pre-trained segmentation model. Notably, the simplicity of the scenes allows for direct use of ground truth segmentation, as observed in PLATO~\cite{piloto2022intuitive}. For each frame, the perception module employs a Component Variational Autoencoder (Component VAE)~\cite{burgess2019monet} to transform each input image into a concealed vector representation $\langle z^0_{1:K}, z^1_{1:K}, ..., z^T_{1:K} \rangle$, where $K$ represents the object count per frame.

\paragraph{Reasoning}

The reasoning module leverages the object embeddings obtained from the perception module as input and endeavors to enhance scene comprehension by inferring the attributes of occluded objects, whose masks remain vacant due to occlusion. This aspect employs two Transformer models to refine object embeddings and recover hidden objects. Both Transformers adopt flattened spatial-temporal embeddings and apply global attention mechanisms to contextualize information. The first Transformer refines input features of occluded objects to align with a learned dynamics module, producing $\tilde{z}$. The second Transformer is responsible for recuperating objects concealed within observation sequences of both original and refined features. It's important to note that object recovery mirrors Masked Autoencoding~\cite{he2022masked}, treating a random object as absent and necessitating reconstruction from contextual cues. Drawing from these observations, we train the second Transformer similarly to Masked Autoencoders (MAE).

\paragraph{Dynamics}

The dynamics module predicts object embeddings $\hat{z}_{1:K}^{t+1}$ in the succeeding frame based on the preceding frame's refined object embeddings $\tilde{z}_{1:K}^{1:t}$. This involves employing the interaction dynamics module introduced in PLATO~\cite{piloto2022intuitive}, supplemented by a residual module. Unlike PLATO, we employ object embeddings subsequent to the reasoning module and jointly train the modules.

\begin{figure*}[t!]
    \centering
    \includegraphics[width=\linewidth]{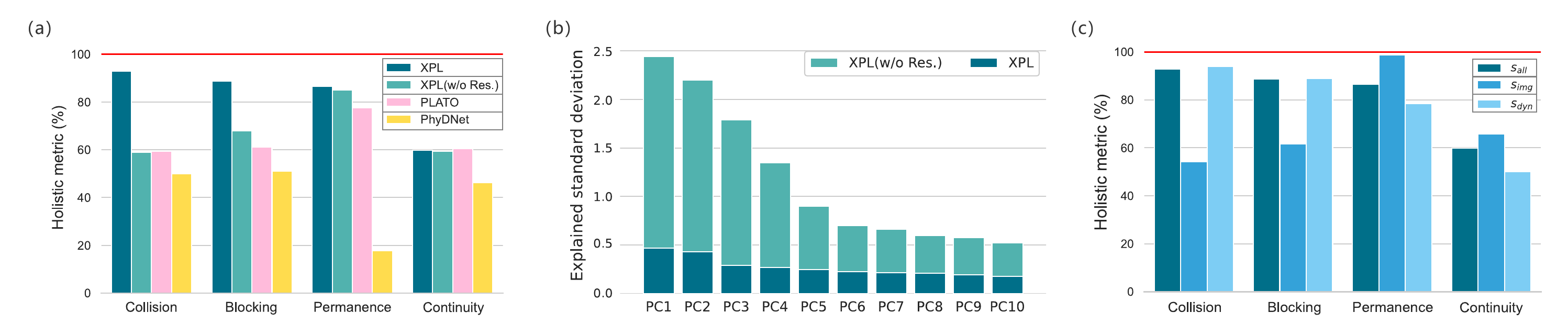}
    \caption{\textbf{(a) Performance of different models on \dataset under the holistic metric.} The red line denotes the ideal performance. \textbf{(b) PCA with or without residual connection.} The first ten principal components are shown. \textbf{(c) Results from each score component.}}
    \label{fig:result}
\end{figure*}

\subsection{Model training}

Initially, we pre-train the perception module to equip the system with foundational visual capabilities. Precisely, the perception module undergoes pre-training using RGBD images and segmentation masks. Throughout this phase, we segment objects and employ masked images for VAE training. During image reconstruction, depth information assists in calculating object mask details.

We then train one Transformer and the dynamics module, with latent codes frozen from the perception module, in an end-to-end manner employing the following loss:
\begin{align}
    \tilde{z}   & = f_{\text{inf}}(z) \nonumber                                               \\
    \mathcal{L} & = \norm{f_{\text{dyn}}(\tilde{z}_{1:K}^{0:t}) - \tilde{z}_{1:K}^{t + 1}}_2,
    \label{eqn:init}
\end{align}
Here, the Transformer employs the architecture featured in Aloe~\cite{ding2021attention} ($f_{\text{inf}}(\cdot)$), while the dynamics prediction module aligns with PLATO~\cite{piloto2022intuitive} ($f_{\text{dyn}}(\cdot)$). The second Transformer is trained independently using MAE.

\section{Experiments}

In this section, we thoroughly evaluate the performance of \ac{method} using our \dataset dataset across different experimental configurations: predicting future phenomena (predictive setup), interpreting existing phenomena (hypothetical setup), and understanding past occurrences given future conditions (explicative setup). We compare \ac{method} against PhyDNet~\cite{guen2020disentangling}, a video prediction model, and PLATO~\cite{piloto2022intuitive} in our \dataset dataset. These models are evaluated under two different metrics.

\subsection{Defining accuracy and surprise}

Before delving into different evaluative configurations, we first introduce how accuracy and surprise are formally defined.

In developmental psychology experiments on \ac{voe}, a surprise was defined by comparing infants' responses to normal scenes with those that violate expectations. Similar to existing works~\cite{smith2019modeling}, we borrow the idea and define the model accuracy as the relative scores between two videos, one that violates intuitive physics laws and another that does not:
\begin{equation}
    \text{Accuracy} = \frac{1}{N}\sum \mathbb{1}[s_\text{nor} < s_\text{sur}],
    \label{eqn:accuracy}
\end{equation}
where $N$ denotes the total number of such pairs, and $s_\text{nor}$ and $s_\text{sur}$ are scores of a normal physics video and one that violates physics, respectively. The scores are computed as the sum of the difference between the inferred results from the observation and that from the dynamics module's prediction, \ie,
\begin{equation}
    s = s_\text{img} + s_\text{dyn},
    \label{eqn:total_score}
\end{equation}
where
\begin{equation}
    s_\text{img} = \sum_{t = 1}^{T} \ell(\text{I}_t, \sum_{i} f_\text{dec}(\tilde{z}^t_i)),
\end{equation}
and
\begin{equation}
    s_\text{dyn} = \sum_{t = 2}^{T} \ell(\sum_{i} f_\text{dec}(\tilde{z}^t_{i}), f_\text{dec}(f_\text{dyn}(\tilde{z}^{0:t-1}_{1:K})).
\end{equation}
Here, $f_\text{dec}(\cdot)$ denotes the learned decoder in our VAE, and we use MSE loss for $\ell(\cdot)$.

\begin{figure*}[t!]
    \centering
    \includegraphics[width=\linewidth]{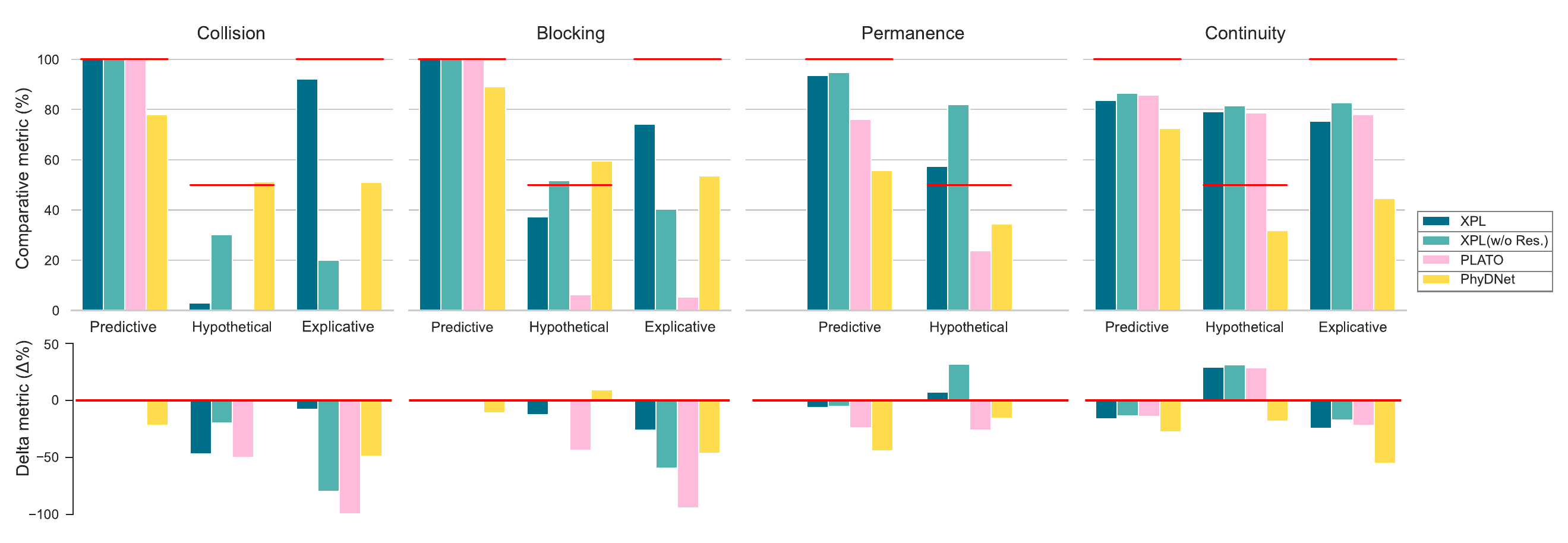}
    \caption{\textbf{Performance of different models on \dataset under the comparative metric.} The red line denotes the ideal performance. The top part shows the absolute comparative values and the bottom part shows the difference from the ideal.}
    \label{fig:comparison}
\end{figure*}

\subsection{The holistic metric}

Similar to Smith \etal~\cite{smith2019modeling}, we adopt the holistic metric to evaluate \ac{voe} effects in all pairs of unexpected and normal event videos. Ideally, an intuitive physics model should produce higher surprise scores for unexpected events. Formally, the holistic metric is defined as such,
\begin{equation}
    \frac{1}{n_s n_c}\sum_{i,j}\mathbf{1}[s(x_i^{+}) > s(x_j^{-})],
    \label{eqn:holistic_metric}
\end{equation}
where $x_i^{+}$ and $x_j^{-}$ denote the unexpected and normal videos and $n_s$ and $n_c$ are the number of unexpected and normal videos. This metric aggregates results from all confounding factors, including interference from colors, shapes, scene complexity, \etc. Therefore, it provides a holistic view of models' understanding of intuitive physics events; models need to judge the unexpectedness of outcomes from the intuitive physics perspective, disentangling all other confounding factors.

As shown in \cref{fig:result} (a), we measure the holistic value on different models on \dataset. Both \ac{method} and PLATO show better performance in all four testing scenarios, though with a notable gap from perfection. \ac{method} is significantly better than PLATO in the collision, blocking, and permanence, but less so in continuity. We also compare different dynamic modules, with or without residual, in \ac{method}. The results show that the residual connection in the dynamics module plays a critical role in our system, as evidenced by results for collision and blocking. An in-depth analysis from Principal Component Analysis (PCA) in \cref{fig:result} (b) shows that after adding the residual connection, the standard deviation in different principal components is particularly reduced, making learning easier.

To investigate the contribution of each of the two surprise components in \cref{eqn:total_score}, we compute the holistic metric from each of them separately. As shown in \cref{fig:result} (c), the performance of $s_\text{dyn}$ is superior to that of $s_\text{img}$ in the collision and blocking scenarios, whereas the performance of $s_\text{img}$ is better in permanence and continuity. This result implies that the violation of physical knowledge plays a more important role in collision and blocking. In contrast, the mismatch from the observation is a more crucial factor for permanence and continuity. Thus, the residuals in \ac{method}, explicitly taking earlier information into computation, could exert a greater influence on the dynamic module and its impact in the collision and blocking scenarios as shown in \cref{fig:result} (a).

The holistic metric only provides a global view of how a model understands intuitive physics. To paint a more complete landscape of a model, we look deeper into the comparative metric in the next section.

\subsection{The comparative metric}

The comparative metric, similar to ones proposed in literature~\cite{riochet2020intphys,weihs2022benchmarking}, is calculated in a pair of the unexpected and normal events within one specific setting in each scenario,
\begin{equation}
    \frac{1}{n}\sum_{i}\mathbf{1}[s(x_i^{+}) > s(x_i^{-})],
\end{equation}
where $x_i^{+}$ and $x_i^{-}$ are the two paired videos in each settings and $n$ is the number of such pairs. The comparative metric is also most commonly used in evaluating infants' intuitive physics knowledge in developmental psychology~\cite{baillargeon1985object,lin2020infants}.

Whereas the holistic metric describes whether an observation sequence is absolutely surprising from a holistic perspective, the comparative metric assesses whether one observation sequence is more surprising than another from a comparative perspective. Although the holistic metric provides an overall perspective, it lacks the detailed results of the three specific cases the comparative metric provides; see \cref{fig:explain}. In each scenario in \dataset, the two videos in the hypothetical setting are likely to occur, while only one of the two videos in the predictive and explicative settings is likely to occur. Therefore, the comparative metric in the hypothetical setting should be ideally 50\%, while the metric in the predictive and explicative settings should be ideally 100\%.

\cref{fig:comparison} shows the comparative values of different models. The results in the predictive setting indicate that current AI systems, even as simple as general video prediction, can easily predict future outcomes accurately for such a simple task. However, when it comes to the setting that requires reasoning and explanation (\ie, explicative), only \ac{method} can consistently achieve over 50\%. When common predictive models can only predict future occurrences based on past conditions, \ac{method} can reason about the past conditions that lead to the observation, a critical ability necessary for successfully solving the explicative setting.

\begin{figure*}[t!]
    \centering
    \includegraphics[width=\linewidth]{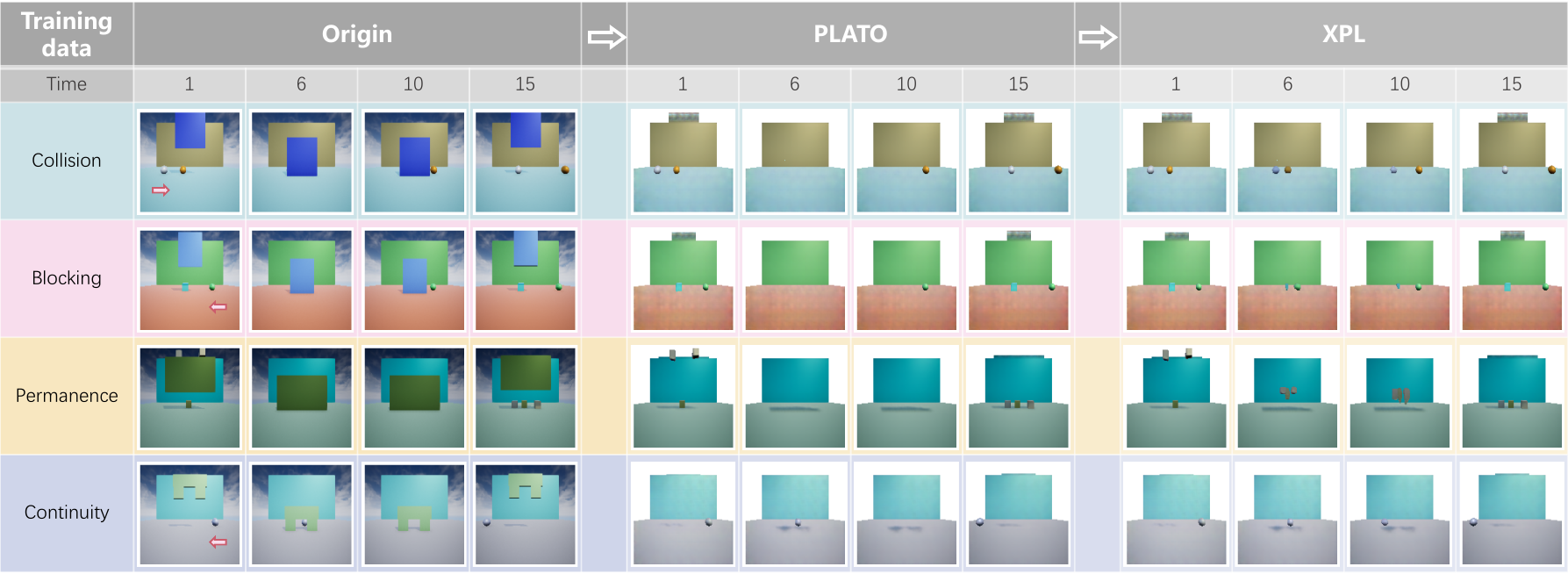}
    \caption{\textbf{Training: Visualization of the internal representation in PLATO and \ac{method} during training.}}
    \label{fig::train_viz}
\end{figure*}

Of these, the hypothetical setting is where we notice the most performance volatility. For the hypothetical setting, both a random-answering human subject and an ideal human subject with perfect understanding would reach 50\% accuracy. However, this is exactly why this problem is intriguing for psychologists. From this perspective, a model achieving 50\% could mean it is either the worst or best. While in the hypothetical setup, PhyDNet achieves nearly 50\%, it can only reach random-level performance in the explicative setting, showing that the model does not understand different possibilities behind the wall.
This is why the explicative setting is so important. The explicative setting provides more new information in the video follow-up than the hypothetical setting. As shown in \cref{fig:explain}, the new information will change a possible scene to an impossible scene in the hypothetical setting. The metric gap between the hypothetical setting and explicative setting shows the power of the explanatory abilities. \ac{method} demonstrates this property on both collision and blocking scenarios, especially on the collision scenario, where this gap reaches close to 90\%.

Although the \ac{method} with or without a residual module both have the reasoning module, they still have different explanatory abilities for hypothetical and explicative settings. In collision and blocking tasks, residuals' presence improves the explicative but not the hypothetical setting. The residual module enhances the connection between two consecutive frames, allowing the reasoning module to better infer the previous state based on the subsequent state. The main difference between the hypothetical and explicative setting is the inclusion of follow-up information. In the explicative setting, the presence of follow-up information enhances the performance of the reasoning module (with residual module) due to more subsequent state information. However, in the hypothetical setting, the absence of follow-up information negatively impacts the module's performance.

Overall, \ac{method} improves over previous state-of-the-art but still fares worse on collision and continuity. While developmental psychology experiments have found the ability in infants~\cite{aguiar2002developments}, it remains a challenge for AI systems.

\subsection{Visualization results}

\begin{figure}[t!]
    \centering
    \includegraphics[width=\linewidth]{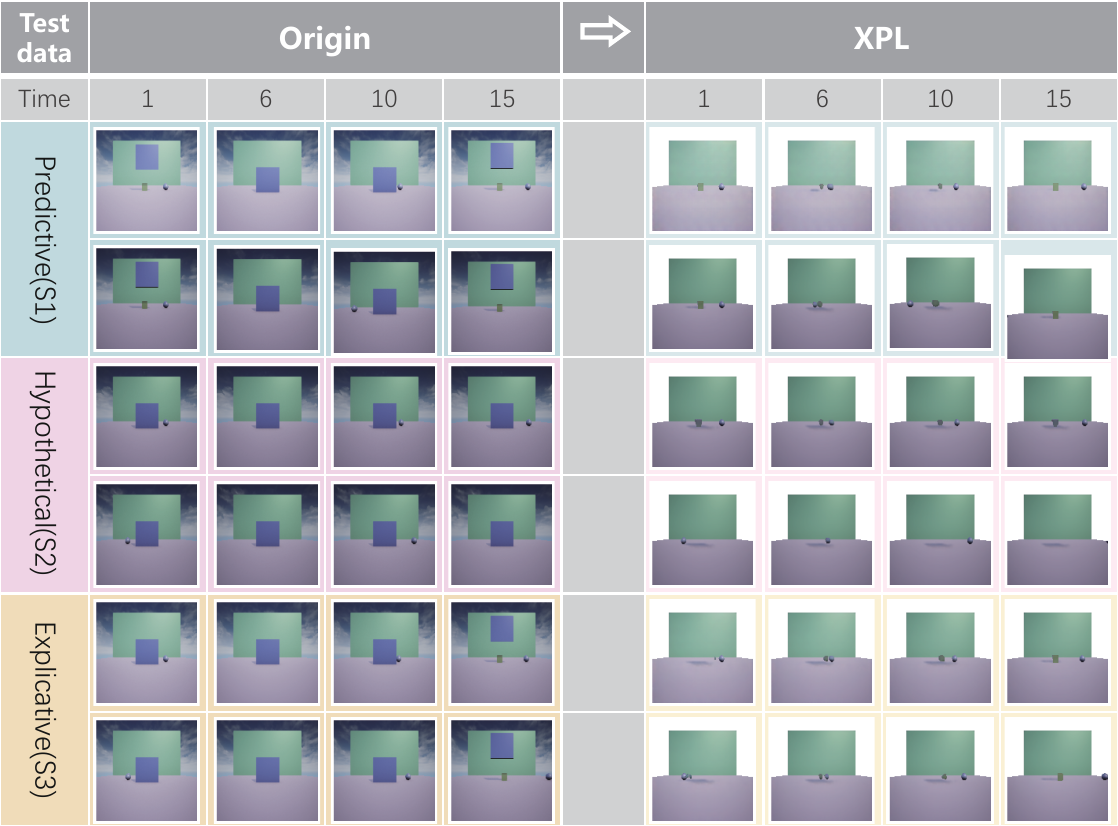}
    \caption{\textbf{Testing: Visualization of the inferred internal representation in \ac{method} during testing.} This example corresponds to the settings in \Cref{fig:explain}.}
    \label{fig:test_viz}
\end{figure}

The challenge of visual occlusion persists in computer vision. Unless the ground-truth value is given directly, it is difficult to characterize occluded objects by vision alone, especially in the case of complete occlusion. However, humans can deduce occluded objects and corresponding physical phenomena intuitively, even under complete occlusion. We investigate whether \ac{method} can reason about occluded objects through visualization.

We visualize occluded objects within the learned representation. Specifically, we mask the token associated with the wall and decode the resulting features to assess the model's ability to reconstruct hidden objects. Training visualization results are presented in \cref{fig::train_viz}. Notably, PLATO lacks a dedicated reasoning module for occluded objects, resulting in an inability to recover occluded factors. Conversely, \ac{method} gradually learns to infer the presence of occluded objects behind the wall to explain observations. Crucially, we never provide ground-truth occluded object representations during training, emphasizing the importance of synchronized training of the inference and dynamic modules. This approach allows \ac{method} to achieve improved occluded object restoration, though it still falls short of ground-truth results (\cref{fig::train_viz}).

For test visualization, detailed results corresponding to \cref{fig:explain} are showcased in \cref{fig:test_viz}. The predictive setting demonstrates \ac{method}'s accurate reconstruction of observed objects. In the hypothetical setting, \ac{method} provides coherent explanations involving hidden object interactions. In the explicative setting, the occluder is lifted toward the end of the videos, resulting in surprising outcomes.

To conclude, \ac{method} proficiently reconstructs occluded objects and provides visual explanations for various events, underscoring its capacity to reason about hidden factors in the context of intuitive physics.

\section{Conclusion and discussion}

In this paper, we introduced \dataset, a novel explanation-based \acf{voe} dataset consisting of four distinct scenarios, each encompassing three unique settings: predictive, hypothetical, and explicative. While the predictive setting aligns with conventional \ac{voe} tasks, the other two settings focus on evaluating a model's explanatory capacity. Our proposed \ac{method} combines reasoning and explanation processes to address occluded objects, offering enhanced performance within the \dataset settings. Our experiments revealed that \ac{method} excels in scenarios requiring explicit explanations for occluded objects, positioning it ahead of other methodologies. Notably, the decoded representation from \ac{method} offers visual explanations for occluded events, highlighting its ability to reason about hidden factors.

Our work underscores the pivotal role of explanations in \ac{voe} tasks, particularly concerning occluded objects and their contribution to video comprehension. Even when objects are obscured by walls, the possibility of underlying physical events remains, and a model equipped with explanation capabilities performs more adeptly in such situations. The capacity to reason about occluded objects extends the model's scope beyond mere video prediction, enabling it to capture intuitive physics principles more effectively.

However, certain challenges persist. Notably, \ac{method} encounters difficulties in scenarios that demand high-level explanations, such as the hypothetical setting in collision or continuity (\cref{fig:comparison}). These limitations underscore the need for further advancements in the reasoning aspect of our model, paving the way for future research. The ability to handle complex interactions and provide meaningful explanations remains a challenging aspect that requires careful consideration in model design.

In conclusion, while our model's reasoning capabilities are still a work in progress, our study sheds light on the integration of explanations into \ac{voe} tasks, aiming to develop models with a level of intuitive physics comprehension akin to infants. The focus on occluded objects and their explanatory potential broadens the scope of \ac{voe} tasks and encourages the development of AI systems with deeper understanding.

\subsection{Limitations}

\paragraph{Method} Despite its strengths, \ac{method} faces certain limitations. It struggles in some experiments, particularly the hypothetical setting in collision or continuity (\cref{fig:comparison}), where its performance falls short of human-like comprehension. Furthermore, our explanation process employs a basic Transformer module, lacking physics-related inductive biases that could enhance performance. A promising direction for future research lies in incorporating domain-specific inductive biases that exploit physical principles to improve reasoning and explanatory capabilities.

\paragraph{Accuracy metric} Although our accuracy metrics draw inspiration from developmental psychology experiments and prior works, they rely on video comparisons to evaluate violations of intuitive physics. This approach, while effective, assumes that one of the videos violates intuitive physics laws, even if the difference in surprise values is marginal. As a result, the method might struggle to achieve the desired metrics in scenarios like the hypothetical setting. Exploring metrics that focus on higher-level concepts and the detection of fundamental violations could yield insights into the underlying mechanisms that drive these evaluations.

\paragraph{Dataset} \dataset pioneers the evaluation of physical explanatory abilities in \ac{voe} tasks. However, our test scenarios could be more diverse and comprehensive. Future efforts will expand and diversify these scenarios to create a more robust framework for testing intuitive physics understanding in \ac{voe}. By incorporating a wider range of physical phenomena and interactions, future datasets can challenge AI systems with greater complexity.

\subsection{Future Directions}

Future research should focus on refining \ac{method}'s reasoning capabilities, enhancing its performance in scenarios demanding higher-order explanations. Introducing more sophisticated physics-based inductive biases could contribute to better occluded object reasoning. Additionally, exploring hybrid approaches that combine neural networks with symbolic reasoning could lead to more advanced models with enhanced explanatory capabilities.

Additionally, \dataset can serve as a stepping stone for designing more intricate and varied \ac{voe} scenarios. Incorporating more complex physical interactions, occlusions, and multiple objects would lead to a richer and more challenging testbed for evaluating AI systems' intuitive physics comprehension. Diverse scenarios can provide comprehensive evaluation of models' understanding across a wide range of intuitive physics principles.

In summary, our study provides insights into the integration of explanations in \ac{voe} tasks and sets the stage for future advancements in both model design and dataset development. The intersection of explanations and intuitive physics comprehension holds promise for creating AI systems that not only predict events but also understand the underlying physical principles that govern them.

\section*{Acknowledgment}

The authors would like to thank four anonymous reviews for constructive feedback, Huiyin Li (BIGAI) for designing the figures, and NVIDIA for their generous support of GPUs and hardware. This work is supported in part by the National Key R\&D Program of China (2022ZD0114900) and the Beijing Nova Program.

    {
        \small
        \bibliographystyle{config/ieee_fullname}
        \bibliography{reference_header,reference}
    }

\clearpage
\appendix
\renewcommand\thefigure{A\arabic{figure}}
\setcounter{figure}{0}
\renewcommand\thetable{A\arabic{table}}
\setcounter{table}{0}
\renewcommand\theequation{A\arabic{equation}}
\setcounter{equation}{0}
\pagenumbering{arabic}
\renewcommand*{\thepage}{\arabic{page}}
\setcounter{footnote}{0}

\section{Dataset}

\subsection{Test data}

For the VoE task, we divided the four scenarios into 11 groups, each with two comparison cases. The setups in the testing data are very similar to the ones in the training data except for the behavior of the wall. All scenarios except Permanence contain predictive, hypothetical, and explicative settings. The predictive and explicative settings contain both plausible and implausible events, while the hypothetical setting contains two plausible events. In the predictive setting, the wall is moved away at the beginning and end of the video, so all information is shown at the beginning and end of the video. In the hypothetical setting, the wall always stays in the middle of the scene. In the explicative setting, the wall is moved away only at the end of the video, so new information is shown to the model at the end of the video.

\paragraph{Collision}

\begin{figure*}[t!]
    \centering
    \includegraphics[width=\linewidth]{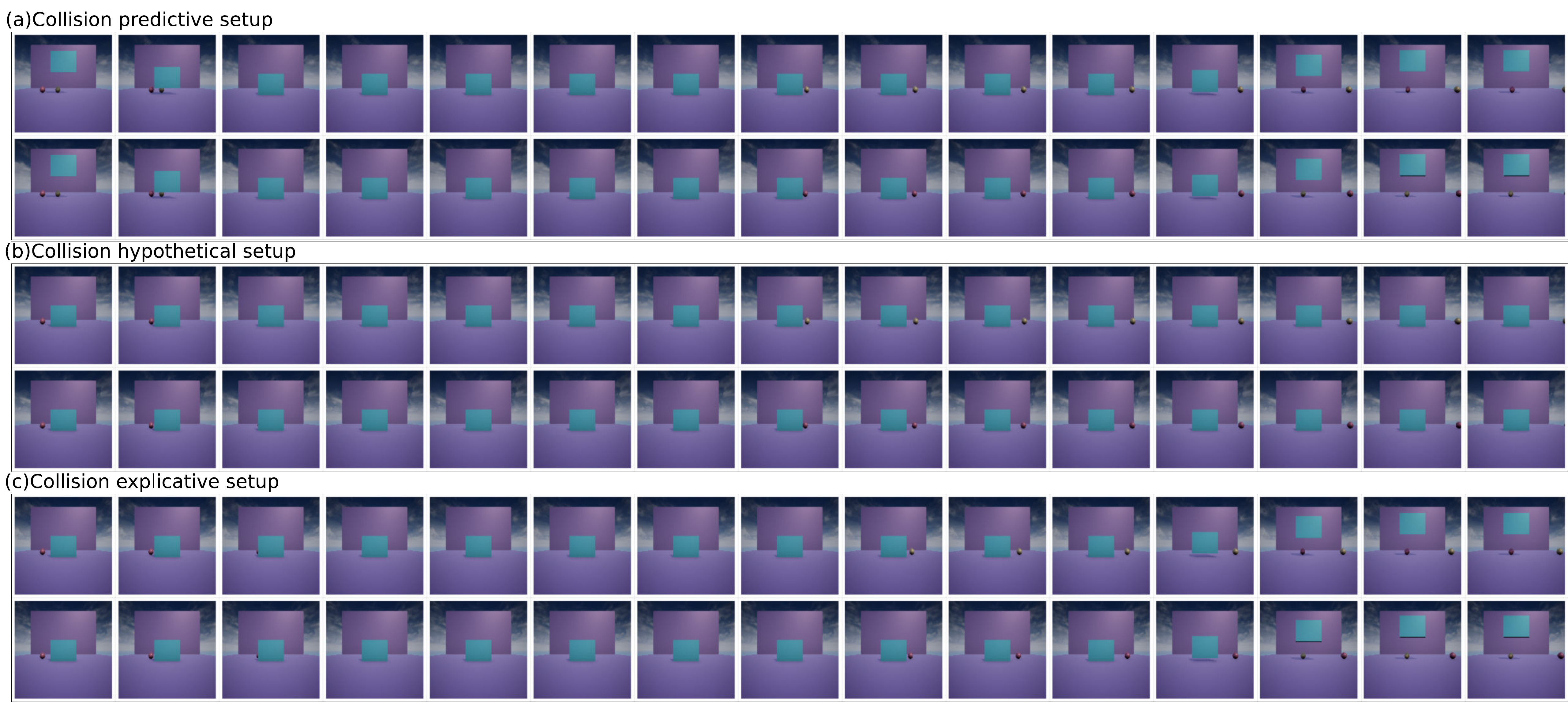}
    \caption{Collision test groups.}
    \label{fig_supp:test_collision}
\end{figure*}

The Collision scenario is shown in \cref{fig_supp:test_collision}. Collision contains predictive, hypothetical, and explicative settings. In the predictive setting, the wall is moved away at the beginning and end of the video, so two balls are visible to the model. We can easily tell from intuitive physics that the case in the first row is possible while the case in the second row is not, because the red ball cannot pass through the blue ball without collision. In the hypothetical setting, the wall always stays in the middle of the scene, so we can not tell how many balls there are in the scene. As we can not infer if a blue ball is hidden behind the wall at the beginning of the video, both cases in the setting are possible. In the explicative setting, the wall is moved away at the end of the video, so additional information is given. We can infer that a blue ball must be hidden behind the wall, so the case in the first row is possible, while the case in the second row is not.

\paragraph{Blocking}

\begin{figure*}[t!]
    \centering
    \includegraphics[width=\linewidth]{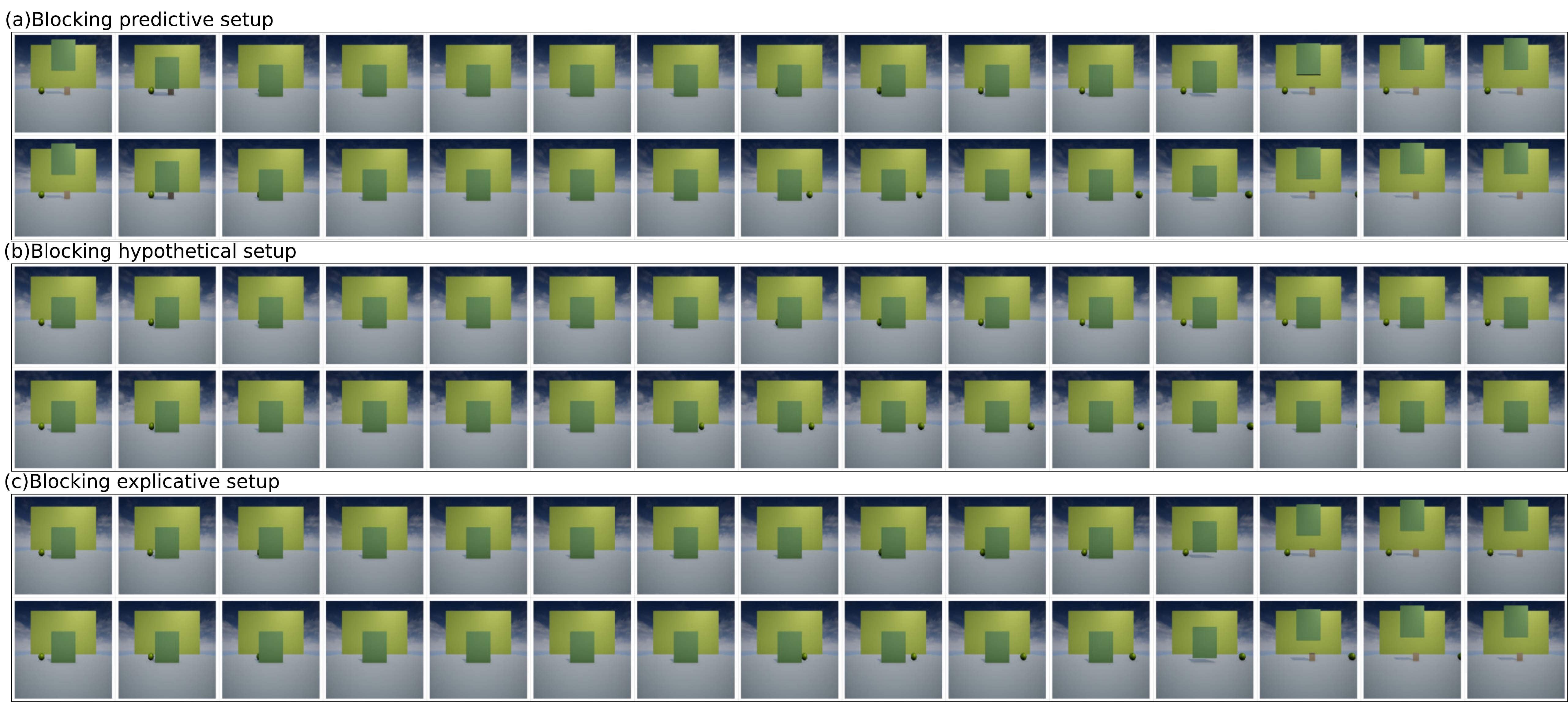}
    \caption{Blocking test groups.}
    \label{fig_supp:test_block}
\end{figure*}

The Blocking scenario is shown in \cref{fig_supp:test_block}. The Blocking scenarios are similar to the Collision scenarios, except that the ball hidden behind the wall is replaced by a fixed cube. In the predictive setting, the wall is moved away at the beginning and end of the video, so the cube is visible to the model. Similar to Collision, we can easily tell that the case in the first row is possible while the case in the second row is not, because the blue ball can not pass through the green cube without collision.  In the hypothetical setting, the wall always stays in the middle of the scene, so we can not tell if there is a cube behind the wall. Therefore, both cases in the setting are possible. In the explicative setting, the wall is moved away at the end of the video, so we can infer that a cube must be hidden behind the wall. Furthermore, we can tell that the case in the first row is possible while the case in the second row is not.

\paragraph{Permanence}

\begin{figure*}[t!]
    \centering
    \includegraphics[width=\linewidth]{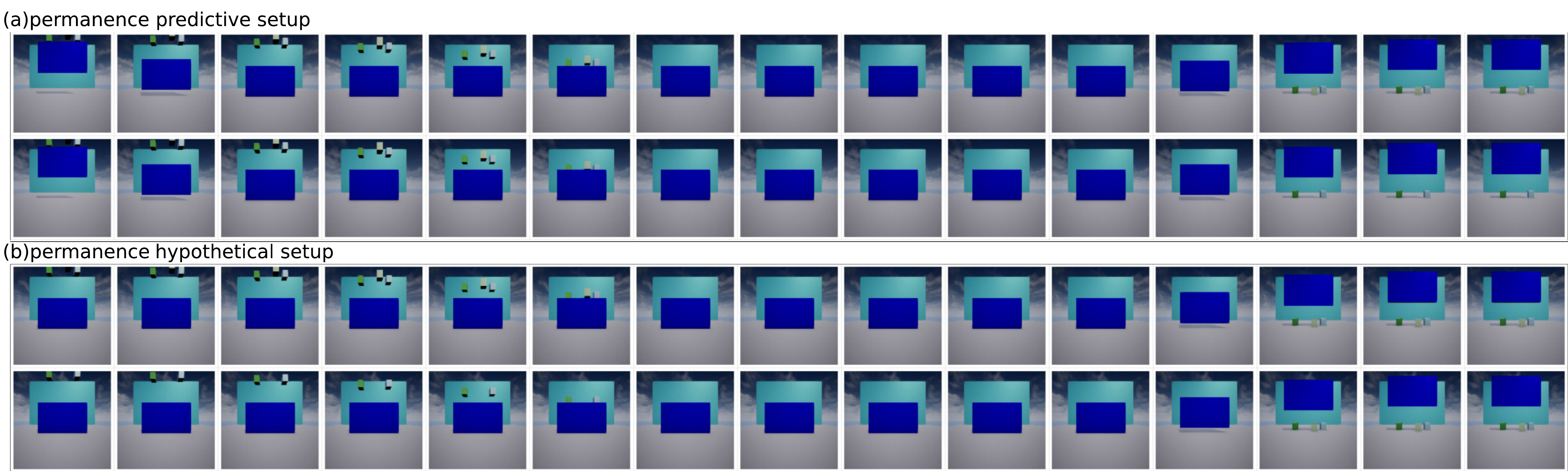}
    \caption{Permanence test groups.}
    \label{fig_supp:test_permanence}
\end{figure*}

The Permanence scenario is shown in \cref{fig_supp:test_permanence}. In the Permanence scenarios, three cubes are randomly divided into two groups (allowing empty groups), where cubes in the first group are dropped to the ground and the second rest on the floor. We do not have an explicative setting for this scenario, as there is no new evidence at the end of the video. In the predictive setting, the wall is moved away at the beginning of the video, so we can infer that there is no object on the ground at the beginning. So the case in the second row is impossible, while the case in the first row is possible. In the hypothetical setting, the wall stays in the middle of the scene at the beginning, so we can not tell if there are cubes on the ground at the beginning, so both cases are possible.

\paragraph{Continuity}

\begin{figure*}[t!]
    \centering
    \includegraphics[width=\linewidth]{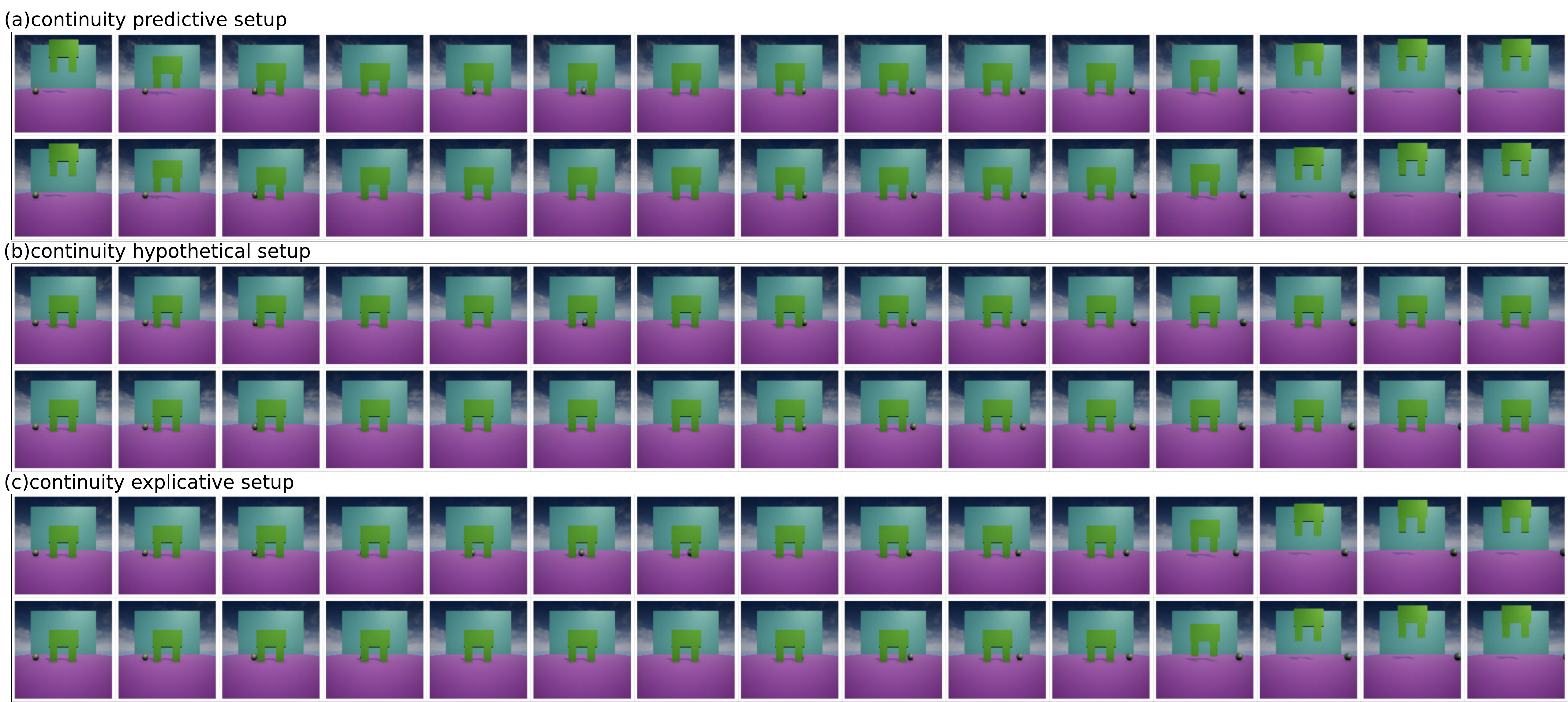}
    \caption{Continuity test groups.}
    \label{fig_supp:test_continuity}
\end{figure*}

The Continuity scenario is shown in \cref{fig_supp:test_continuity}. In the Continuity scenarios, we create a window on the lower half of the wall. In the case of the wall, the ball rolls across the scene. When the ball passes through the wall, it can be seen going from one side to the other. In the predictive setting, the wall is moved away at the beginning of the video, so we can infer that only one ball is in the scene. We can tell that the case in the second row is impossible while the case in the first row is possible. In the hypothetical setting, the wall always stays in the middle of the scene, and we can easily infer that the case in the first row is possible. Considering the case in the second row, we can not tell if there are two balls with the same appearance in the scene, one of which is visible at the beginning and the other one is hidden by the right part of the wall. If that is true, the case in the second row is also possible. So both cases are possible. In the explicative setting, the wall is moved away at the end of the video, so we can infer that there is only one ball in the scene. Thus we can tell that the case in the first row is possible while the case in the second row is not.

\subsection{Train data}\label{sec_sup:train_data}

For four scenarios, we created 5 groups for training. Each of Permanence and Continuity contains 1 group, while Collision and Blocking in total contain 3 groups. Each group contains 2 kinds of cases: cases with a wall and ones without a wall. In the case with a wall, a movable wall stands in the middle of the scene and will be moved away at the beginning and the end of the video. In the case without the wall, everything stays the same except that the wall does not exist, showing that the wall won't interact with other objects physically. Each row in the \cref{fig:train} corresponds to one sampled video in a specific case. See \cref{fig:train} for all training groups.

\paragraph{Control group}

In the control group, a ball rolls across the scene without interacting with other objects, indicating that the environment follows basic physics.

\paragraph{Collision group}

A ball rolls across the scene in the Collision scenario with the wall. Another ball with the same mass but a different color is hidden behind the wall and will collide with the incoming ball, causing the first ball to stop and itself to pass through. In a setting without a wall, the second ball will always be visible.

\paragraph{Blocking group}

The Blocking scenarios are similar to the Collision scenario, except that the ball hidden behind the wall is replaced by a fixed cube. A ball rolls across the scene in the blocking setting with the wall. A fixed cube is hidden behind the wall and will collide with the incoming ball, causing the incoming ball to turn around. In the setting without a wall, everything stays the same except that the wall doesn't exist, and the cube will always be visible.

\paragraph{Permanence group}

In the Permanence scenario, three cubes are randomly divided into two groups (allowing empty groups), where cubes in the first group are dropped to the ground and the second rest on the floor. In the setting with the wall, the wall will be moved away at the end of the video, showing that all of the cubes still exist. In the setting without the wall, the cubes will always be visible.

\paragraph{Continuity group}

In the Continuity scenario, we create a window on the lower half of the wall. In the setting with the wall, the ball rolls across the scene. When the ball passes through the wall, it can be seen going from one side to the other, especially visible from the window. In the setting without the wall, the ball will always be visible.

\subsection{Environment}

Our \dataset dataset comprises 22K+100K procedurally generated scenes using Unreal Engine 4. In addition to the floors and the backgrounds, there are four different object types: balls, cubes, walls, and windowed walls. In all videos, the size of the ball and the cube are the same, while the size of the wall with or without windows are randomly different. The positions of objects are randomly set in the videos, except for the walls in the permanent scenes in which the wall is placed in the middle. All objects, including the floor and the background, are randomly set in different colors.

\begin{table*}[ht!]
    \centering
    \small
    \caption{Spatial broadcast decoder architecture (from top to down).}
    \label{tab_supp:decoder}
    \begin{tabular}{llll}
        \toprule
        Type               & Size    & Activation                 & Comment                                  \\
        \midrule
        Spatial Broadcast  & 8 × 8   & -                          & -                                        \\
        Position Embedding & -       & -                          & -                                        \\
        Conv 5 × 5         & 64      & ReLU                       & stride: 2                                \\
        Conv 5 × 5         & 64      & ReLU                       & stride: 2                                \\
        Conv 5 × 5         & 64      & ReLU                       & stride: 2                                \\
        Conv 5 × 5         & 64      & ReLU                       & stride: 2                                \\
        Conv 5 × 5         & 64      & ReLU                       & stride: 1                                \\
        Conv 3 × 3         & 4       & -                          & stride: 1                                \\
        Channels           & RGBD(4) & Softmax (on depth channel) & softmax(depth × abs($\theta$) × -1000.0) \\
        \bottomrule
    \end{tabular}
\end{table*}

\begin{table*}[ht!]
    \centering
    \small
    \caption{The Transformer architecture (from top to down). The [M] is a learnable mask token for Transformer.}
    \label{tab_supp:fast_reasoning}
    \begin{tabular}{llll}
        \toprule
        Type               & Size           & Activation                 & Comment                                 \\
        \midrule
        LP (1)             & 256            & -                          & -                                       \\
        Mask (1)           & -              & × mask + [M] × (1-mask)    & mask : (size F × N × 1), (value 0 or 1) \\
        Position Embedding & -              & -                          & -                                       \\
        Transformer        & 256, 256 (MLP) & ReLU (MLP)                 & head=8,key=32,layers=6                  \\
        LP (2)             & 256            & -                          & -                                       \\
        Mask (2)           & -              & × (1-mask) + inputs × mask & mask : (size F × N × 1), (value 0 or 1) \\
        \bottomrule
    \end{tabular}
\end{table*}

\section{Model}

\subsection{Perception}

The perception module in \ac{method} is similar to that of Component Variational Autoencoder (ComponentVAE) in the PLATO model~\cite{piloto2022intuitive}. For each object $k$ in an image, we take as input a 128 × 128 RGBD (0-255 for each channel) image $x_k$ that is masked except around the object. Then we use a Vision Transformer~\cite{dosovitskiy2020image} encoder $\Phi$ to encode the image with only one object into a 32-dimensional Gaussian posterior distribution $q_{\Phi}(z_k|x_k)$. The sample from this distribution, $z_k$, is decoded by a spatial broadcast decoder~\cite{watters2019spatial} to an RGBD image. To address occlusion, we use the depth of the decoder image to combine all objects in the image by multiplying them with softmaxed depth values. We first pretrained the perception module by optimizing the variational objective defined in~\cite{burgess2019monet}. We set $\sigma$ to 0.05, $\beta$ to 0.5, and $\gamma$ to 0 to ensure that the model reconstructs object masks without segmentation information in the loss function.

\paragraph{ViT encoder}

We first reshape the 128 × 128 × 4 images into a sequence of flattened 16 × 16 × 256 patches, followed by a linear layer with 256 dimensions. Next, we add 2D position embeddings and learnable embeddings, flatten, and send them to a Transformer. We use 8 multi-head, 32 key dimensions, 1024 MLP layer dimensions, and 6 Transformer layers for the Transformer model~\cite{vaswani2017attention}. Finally, we use an MLP layer with size [512, 64] and a leaky-ReLU activation function to the Transformer output and obtain 32-dimensional Gaussian posterior distributions for each object.

\paragraph{Spatial broadcast decoder}

Our spatial broadcast decoder is similar to that in~\cite{locatello2020object}. As shown in \cref{tab_supp:decoder}, we use position embeddings and CNN model to decode the object embeddings, where the parameter $\theta$ in the softmax layer is learnable, thus representing the mask in terms of depth.

\subsection{Reasoning}

In the reasoning module, we use two Transformer modules to reason the hidden object which is occluded in some or all of the frames. All objects in a video can be reshaped as F × N × D embeddings, where F is 15 frames, N is 8 objects, and D is 32 dimensions in our work. As shown in \cref{tab_supp:fast_reasoning}, we use a Transformer model to reason the masked objects in video, similar to the self-supervised learning module in Aloe~\cite{ding2021attention}; the parameter [M] in the Mask (1) part is learnable.

\paragraph{First Transformer}

We set the mask to 0 for objects that are temporally occluded in some frames, and 1 for others. As shown in \cref{tab_supp:fast_reasoning}, we can use the Transformer model to reason the new object embeddings whose mask equals 0. We use it in both the training and testing steps to have better object embedding for the whole video.

\paragraph{Second Transformer}

In our test dataset, there may be cases where an object is obscured in all frames. So in the training step, we set the mask to 0 for one random object (including empty object) in all frames. Then we can train the second Transformer model in a self-supervised manner. In the test step, we set the mask to 0 for one object that is not visible in all frames. Then we can reason about the occluded object to explain the whole video.

\subsection{Dynamics}

In fact, the occluded objects are never directly seen for the Transformer model. After the first reasoning module, we obtain reasonable video object embeddings based on experience. In the dynamics module, we predict the value of the incremental change of the object embeddings in the time step by using the same dynamics module from PLATO~\cite{piloto2022intuitive} with the only difference in object dimension used (from 16 to 32). We refer the readers to~\cite{piloto2022intuitive} for architectural details.

\begin{table*}[ht!]
    \centering
    \caption{Training parameters. The pre-processed video features are calculated by the Perception module, which is pre-trained.}
    \label{tab_supp:trainpara}
    \resizebox{\linewidth}{!}{
        \begin{tabular}{lllllll}
            \toprule
            Model                                    & batch size                         & training step & optimizer & learning rate & warm step & delay step \\
            \midrule
            Perception module (in \ac{method},PLATO) & 300 (images)                       & 472000        & Adam      & 0.0004        & 2000      & 100000     \\
            \ac{method}                              & 500 (pre-processed video features) & 32000         & Adam      & 0.0004        & 1000      & 10000      \\
            PLATO                                    & 500 (pre-processed video features) & 32000         & Adam      & 0.0004        & 1000      & 10000      \\
            PhyDNet                                  & 100 (videos)                       & 70000         & Adam      & 0.001         & -         & -          \\
            \bottomrule
        \end{tabular}
    }
\end{table*}

\section{Training}

\subsection{Training detail}

In a scene with occlusion, we cannot get the representation of the occluded object directly by observation. Therefore, we first use the dynamics loss on the object embeddings after the first Transformer to train our first Transformer and dynamics model. Then, we use the object embeddings after the first Transformer to train our second Transformer model. We randomly mask an object throughout the video frame and use the model to predict representations of the objects throughout the video, enabling the model to infer whether there is a fully hidden object in the test task.

\subsection{Training parameters}

We first pre-train the perception module and use it for both PLATO and \ac{method}. Then we train our model \ac{method}, PLATO, and PhyDNet with the parameters shown in \cref{tab_supp:trainpara}.

\subsection{Training steps}

\begin{figure*}[t!]
    \centering
    \includegraphics[width=.8\linewidth]{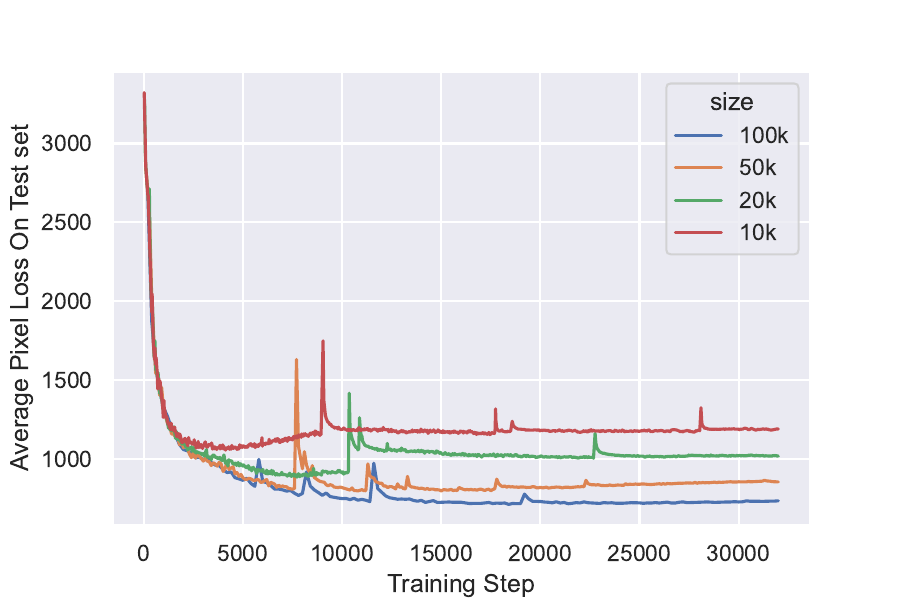}
    \caption{Average pixel loss of test data for different sizes of training data.}
    \label{fig_supp:loss_wrt_size}
\end{figure*}

During the development of the model, we explored how the size of the training dataset impacted the pixel loss of the dynamics module. We use the expected video in the predictive setting of all scenarios as the test dataset to calculate the average pixel loss. \cref{fig_supp:loss_wrt_size} shows that more training data will improve the performance of the dynamics module.

\section{Visualize supplementary}

In the main text, we visualize the reasoning results by our \ac{method} model in the Blocking scenario. Here, we visualize the reasoning results for the rest of the scenarios.

\subsection{Collision}

\begin{figure*}[t!]
    \centering
    \includegraphics[width=.8\linewidth]{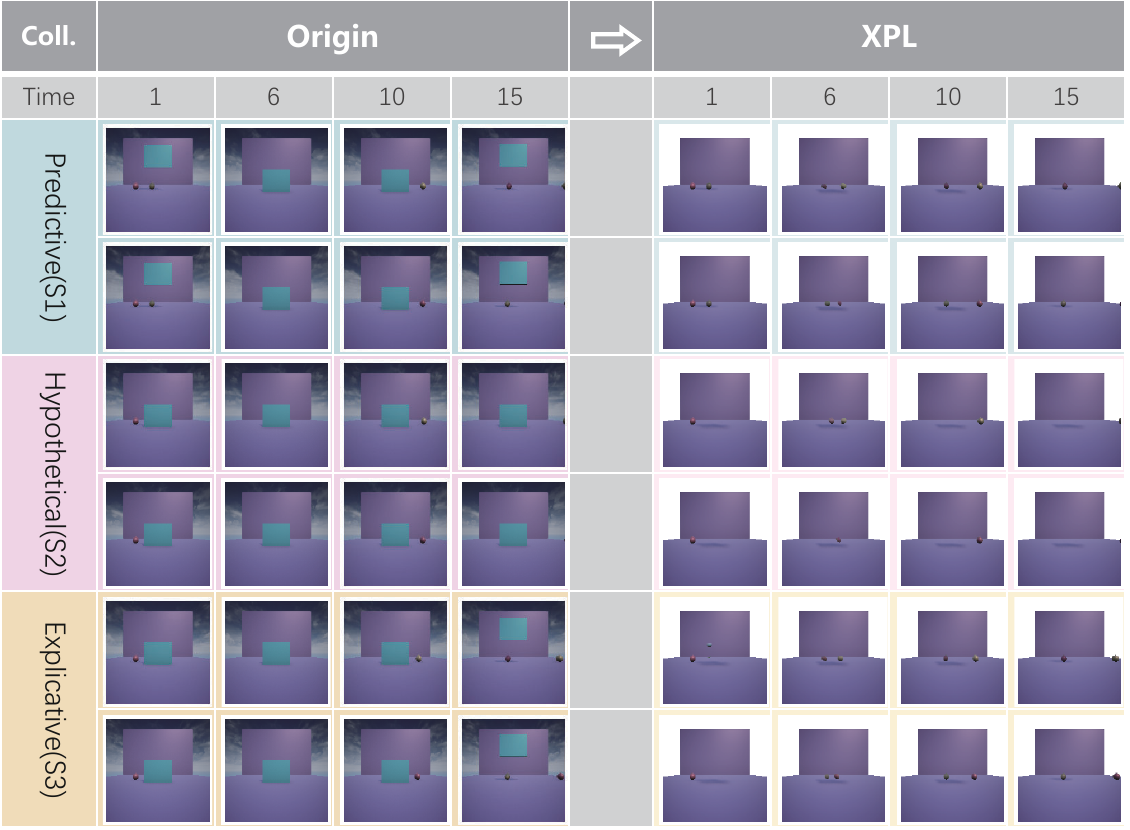}
    \caption{Visualization of the inferred internal representation in \ac{method} during testing in collision scenarios.}
    \label{fig_supp:collision_viz}
\end{figure*}

As shown in \cref{fig_supp:collision_viz}, in the predictive setting, \ac{method} has no problem accurately reconstructing the objects, and the surprise video can be found directly. In the hypothetical setting, the possible explanation for the first video is that another ball collides with the incoming ball. In contrast, no such ball is in the second video, explaining both cases. This result also shows the limitation of our \ac{method} as the incoming ball did not stop behind the wall. In the explicative setting, the occluder is only moved away at the end of the videos. Unlike the hypothetical, when showing a hidden ball behind it, it is impossible for the ball to pass through, causing surprise.

\subsection{Permanence}

\begin{figure*}[t!]
    \centering
    \includegraphics[width=.8\linewidth]{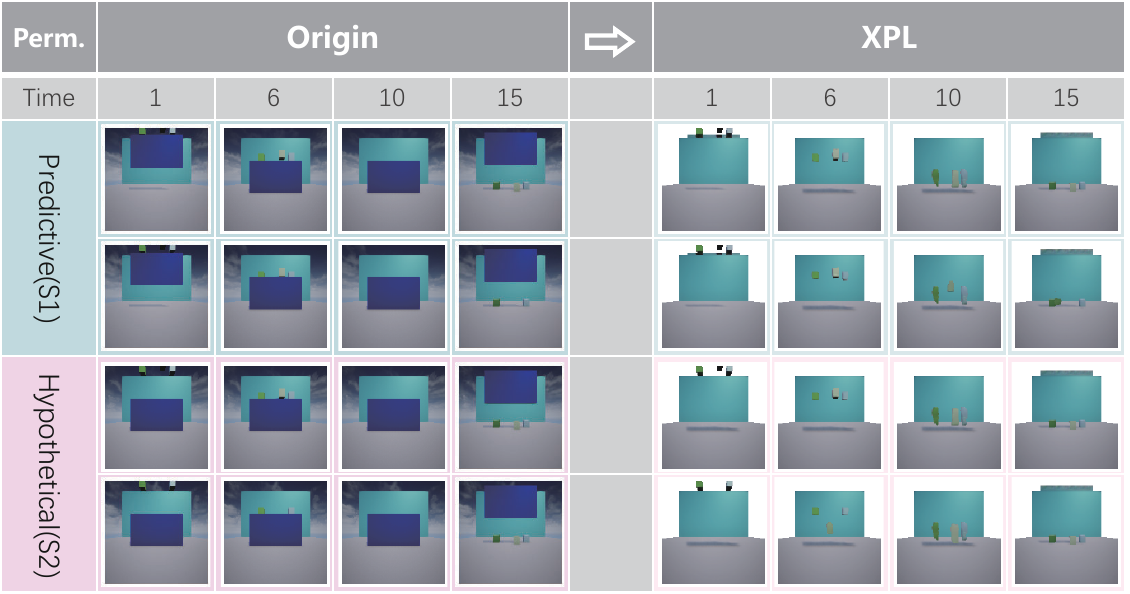}
    \caption{Visualization of the inferred internal representation in \ac{method} during testing in permanence scenarios.}
    \label{fig_supp:permanence_viz}
\end{figure*}

As shown in \cref{fig_supp:permanence_viz}, in the predictive setting, \ac{method} can reconstruct the objects behind the wall, and the surprise video can be found by comparing it with the origin image. The visual effect of the reconstructed objects does not seem to be very well, which is still a limitation of our \ac{method}. In the hypothetical setting, the possible explanation for the second video is that there exists another object behind the wall, and our \ac{method} can reason about the object.

\subsection{Continuity}

\begin{figure*}[t!]
    \centering
    \includegraphics[width=.8\linewidth]{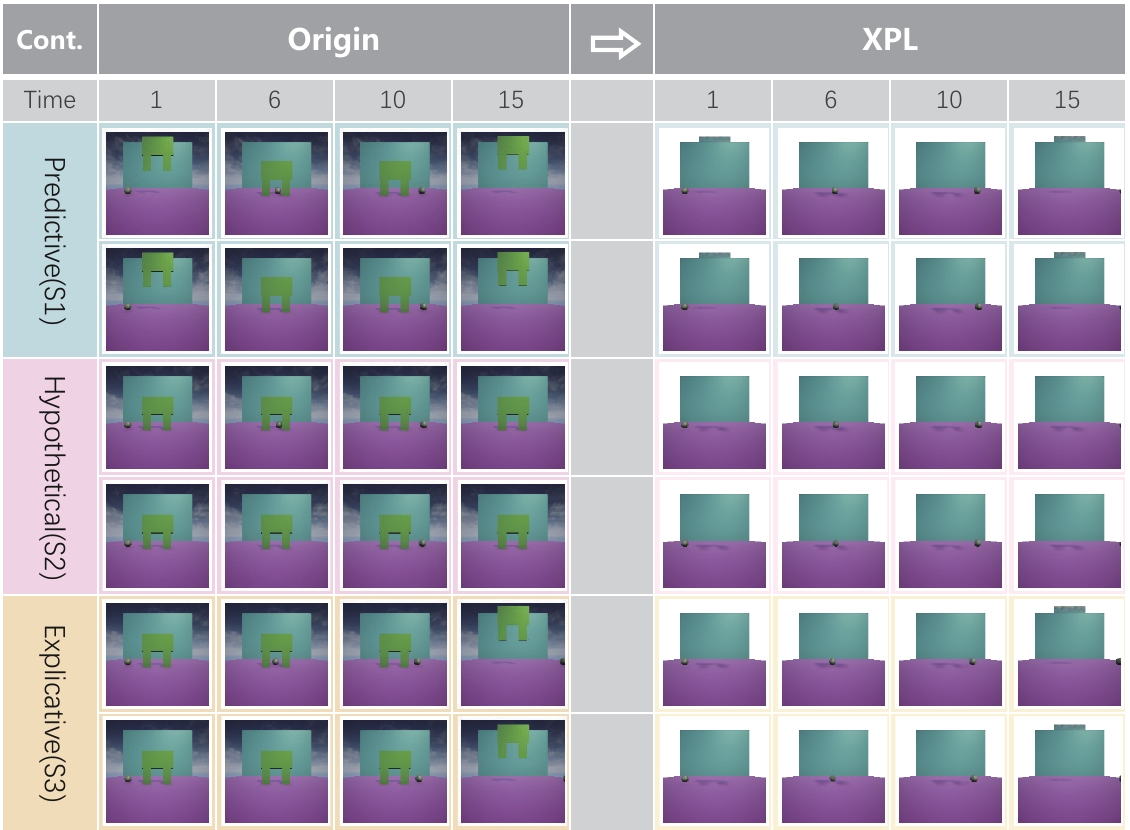}
    \caption{Visualization of the inferred internal representation in \ac{method} during testing in continuity scenarios.}
    \label{fig_supp:continuity_viz}
\end{figure*}

As shown in \cref{fig_supp:continuity_viz}, the visualization results of our \ac{method} are the same in all settings. Even though the visualization results can show surprise in predictive and explicative settings by comparing with the origin videos, our \ac{method} still can not deal with the hypothetical setting due to the limitation discussed in the main text. Our \ac{method} requires given masks and identification of objects. Therefore, it can not reason about the hypothetical setting in continuity by changing the identification of objects and suggesting that there are two same objects as infants do~\cite{aguiar2002developments}.

\end{document}